\pdfoutput=1
\documentclass[sigconf,9pt]{acmart}
\usepackage{makecell}
\usepackage{subcaption}
\usepackage{ulem}
\theoremstyle{remark}
\newtheorem{definition}{Definition}[section]

\newcommand{\Xv}{\boldsymbol{X}}
\newcommand{\Yv}{\boldsymbol{Y}}
\newcommand{\Zv}{\boldsymbol{Z}}

\newcommand{\xv}{\boldsymbol{x}}
\newcommand{\yv}{\boldsymbol{y}}
\newcommand{\zv}{\boldsymbol{z}}
\usepackage{subcaption}
\theoremstyle{remark}

\AtBeginDocument{%
  }

\acmConference[SIGMOD '24]{the 2024
International Conference on Management of Data}{June 9--15,
  2024}{Santiago, AA, Chile}

\acmISBN{978-1-4503-XXXX-X/18/06}

\begin{document}

\title{UniTS: A Universal Time Series Analysis Framework Powered by Self-Supervised Representation Learning}

\author{Zhiyu Liang}
\orcid{0000-0003-0083-2547}
\affiliation{%
	\institution{Harbin Institute of Technology}
	\city{Harbin}
	\country{China}
}
\email{zyliang@hit.edu.cn}

\author{Chen Liang}
\orcid{0000-0002-1093-0362}
\affiliation{%
	\institution{Harbin Institute of Technology}
	\city{Harbin}
	\country{China}
}
\email{1190201818@stu.hit.edu.cn}

\author{Zheng Liang}
\orcid{0000-0003-1844-4366}
\affiliation{%
	\institution{Harbin Institute of Technology}
	\city{Harbin}
	\country{China}
}
\email{lz20@hit.edu.cn}

\author{Hongzhi Wang}
\orcid{0000-0002-7521-2871}
\affiliation{%
	\institution{Harbin Institute of Technology}
	\city{Harbin}
	\country{China}
}
\email{wangzh@hit.edu.cn}\authornote{Corresponding author.}

\author{Bo Zheng}
\orcid{0009-0005-5309-3364}
\affiliation{%
	\institution{CnosDB Inc.}
	\city{Beijing}
	\country{China}
}
\email{harbour.zheng@cnosdb.com}


\renewcommand{\shortauthors}{Zhiyu Liang et al.}

\begin{abstract}
 Machine learning has emerged as a powerful tool for time series analysis. Existing methods are usually customized for different analysis tasks and face challenges in tackling practical problems such as partial labeling and domain shift. To improve the performance and address the practical problems universally, we develop UniTS, a novel framework that incorporates self-supervised representation learning (or pre-training). The components of UniTS are designed using sklearn-like APIs to allow flexible extensions. We demonstrate how users can easily perform an analysis task using the user-friendly GUIs, and show the superior performance of UniTS over the traditional task-specific methods without self-supervised pre-training on five mainstream tasks and two practical settings. 
\end{abstract}




\maketitle

\section{Introduction}

Machine learning methods have achieved high performance in many time series analysis tasks, such as classification, forecasting, and anomaly detection\cite{sktime,tslearn,vldb22-ad-evaluation}. However, existing learning-based time series analysis algorithms still face challenges in real-world scenarios. First, the real-world time series data are usually partially labeled due to the high cost or the lack of knowledge for labeling, while the supervised machine learning techniques require adequate labels to perform well. Second, a common problem in practical applications is the domain shift, i.e., the distributions between the training data samples and the data encountered when deploying the models are different, which makes the models difficult to generalize. Last but not least, while there are many task- and domain-specific approaches, it is an open question to determine appropriate methods for a given application (a.k.a. analysis task and dataset).

\textbf{\underline{Contributions}.}  To deal with the above issues, in this paper, we propose \texttt{UniTS}, a novel universal framework for time series analysis. Our idea is to first perform self-supervised pre-training using the unlabeled data to get unified time series representations, which are more independent of the tasks (e.g. classification or anomaly detection) and domains (i.e., data distributions). Next, the models are learned for arbitrary analysis tasks by appending an output module upon the representations and then fine-tuning the model parameters. \textbf{\textit{Compared to traditional machine learning pipelines that learn end-to-end models from scratch for specific tasks and domains, the \texttt{UniTS} framework has several advantages.}}  

\textbf{\textit{First}}, the pre-training module can leverage the inherent structure of the unlabeled data to learn class-distinguishing information, so that only a few labels are needed for fine-tuning. The additional information learned via self supervision can also improve the performance of different tasks. \textbf{\textit{Second}}, benefiting from the self-supervised representation learning, the pre-training module can learn features transferable across domains by disentangling the domain and class information~\cite{shen2022connect}. \textbf{\textit{Third}}, as \texttt{UniTS} produces unified representations for different pre-training models and downstream analysis tasks, we design a feature fusion module that automatically combines the features of diverse models to jointly facilitate the tasks, so that to avoid the method selection for applications. \textbf{\textit{In addition}}, the \texttt{UniTS} pipeline can be more efficient when performing several tasks on one dataset, because the pre-training is needed only once, while the fine-tuning usually requires a much smaller number of iterations compared to the training from scratch.

\begin{table*}[t]
    \centering
    \caption{Examples of five mainstream time series analysis tasks.}
    \vspace{-3ex}
    \resizebox{.9\linewidth}{!}{
    \begin{tabular}{lcc}
    \toprule
    \textbf{Task}     &  \textbf{Target} & \textbf{Description} \\
    \midrule
    Classification & $y_i \in \{c_j|j=1,\ldots,C\}$ & The class label. \\
    \hline
    Clustering & $y_i \in \{j|j=1,\ldots,C\}$ & The cluster assignment.\\
    \hline
    Forecasting & $\yv_i = \xv_{i,\ T+1:T+H}$ & The values at the $H$ subsequent time steps.\\
    \hline
    Anomaly detection & $\yv_i \in \{0,1\}^{T}$ & \makecell[c]{The bool values indicating whether the observations at each time step is anomalous.}\\
    \hline
    \makecell[l]{Missing value imputation} & $\yv_i = \xv_{i,j,k}, (j,k) \in P_i$ & \makecell[c]{The missing values. $P_i$ is the set of index indicating the positions of the missing values.} \\
    \bottomrule
    \end{tabular}}
    \label{tab:task_examples}
    \vspace{-1ex}
\end{table*}
\begin{figure*}[htbp]
    \centering
    \includegraphics[width=.67\linewidth]{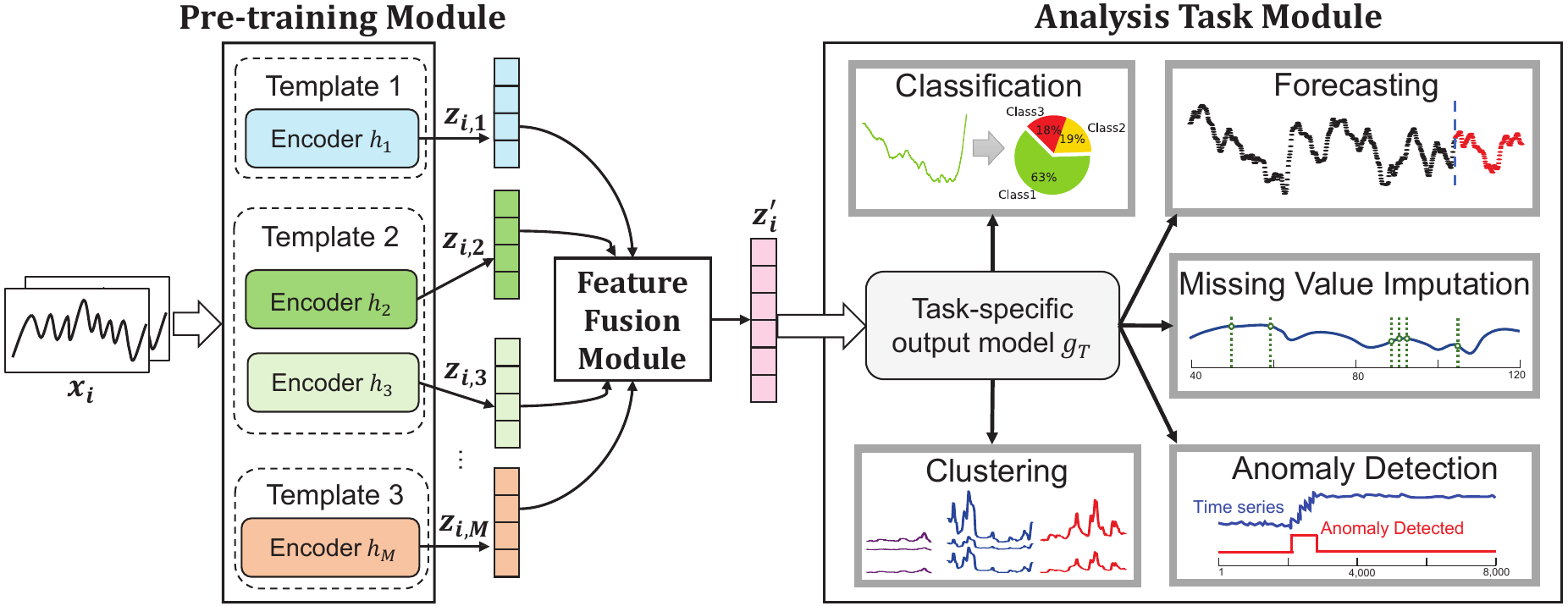}
    \vspace{-1.8ex}
    \caption{Framework overview of \texttt{UniTS}.}
    \label{fig:framework}
    \vspace{-1.2ex}
\end{figure*}
While there exist several machine learning tools for time series, such as tslearn~\cite{tslearn} and sktime~\cite{sktime}, to the best of our knowledge, our \texttt{UniTS} is \textbf{\textit{the first to incorporate self-supervised representation learning for universal time series analysis to achieve the aforementioned benefits}}. We integrate user-friendly Web interfaces and flexible modes of hyper-parameter configuration for usability. \texttt{UniTS} is designed as a framework that is \textbf{\textit{agnostic to model architecture, pre-training algorithm, feature fusion method and analysis task.}} The model architecture is taken as hyper-parameters, and the latter three components are designed as templates using the popular sklearn-like APIs, which allows easy extension to support different models, algorithms and tasks.

In this paper, we demonstrate the usage of \texttt{UniTS} for five mainstream time series analysis tasks, including \textbf{\textit{classification}}, \textbf{\textit{clustering}}, \textbf{\textit{forecasting}}, \textbf{\textit{anomaly detection}}, and \textbf{\textit{missing value imputation}}.  We also examine the performance of \texttt{UniTS} in these tasks and the practical problems of \textbf{\textit{partial labeling}} and \textbf{\textit{domain shift}}. We have made the source code and the supplementary materials publicly available at \textbf{\textit{\url{https://github.com/LceOmlet/UniTS}}} to enable the community to use and extend the system.

\section{System Overview}

\subsection{Problem Formulation}
We first introduce the unified formulation of \textbf{\textit{time series analysis}}, which serves as the basis of our framework.

\begin{definition}[Time Series Analysis]
Given a time series sample $\xv_i = [\xv_{i,1}, \ldots, \xv_{i,T}] \in \mathbb{R}^{D \times T}$ where $\xv_{i,t} \in \mathbb{R}^D$ is the observation at time $t$, $D$ is the number of dimensions, and $T$ is the length of time ranges, a time series analysis task aims to build a function $f$ that can map $\xv_i$ to a task-dependent prediction $\hat{\yv}_i$ (or $\hat{y}_i$), such that the prediction is close to the (usually unknown) target   $\yv_i$ ($y_i$) . Table~\ref{tab:task_examples} shows examples of five important time series tasks.
\end{definition}

 
In machine learning methods, the mapping function $f(\xv_i)$ is learned from a training dataset $\Xv = [\xv_1,\ldots,\xv_N] \in \mathbb{R}^{N \times D \times T}$ and an optional label set $\Yv \in \mathbb{R}^{N \times *}$. \textit{Unlike the traditional approaches that learn $f_{_\mathcal{T}}(\xv_i)$ from scratch for each task $\mathcal{T}$, our \texttt{UniTS} first \textbf{pre-trains} one or more \textbf{task-independent encoders} $h_m$ using only $\Xv$ to map the time series to unified representations, as $\zv_{i,m} = h_m(\xv_i) \in \mathbb{R}^{K}$ ($m=1,\ldots,M$). Then, given a task $\mathcal{T}$, it fuses the features $\zv_{i,1},\ldots,\zv_{i,M}$ to one vector $\zv_i^\prime \in \mathbb{R}^{K^\prime}$ to build a \textbf{task-specific model based on the encoders}, with $f_{_\mathcal{T}}(\xv_i) = g_{_\mathcal{T}}(\zv_i^\prime)$ where $g_{_\mathcal{T}}$ is an output model.} The design goal is to take advantage of various self-supervised pre-training methods to easily achieve universal performance improvement for different analysis tasks and to address the challenges of partial labeling and domain shift.
For the description, we denote $\Zv_m = [\zv_{1,m},\ldots,\zv_{N,m}] \in \mathbb{R}^{N \times K}$.

\subsection{UniTS Framework}
To achieve universal time series analysis via self-supervised pre-training, \texttt{UniTS} is designed with three main modules, including the \textbf{\textit{Pre-training Module}}, the \textbf{\textit{Feature Fusion Module}}, and the \textbf{\textit{Analysis Task Module}}, as illustrated in Figure~\ref{fig:framework}.

First of all, \texttt{UniTS} creates one or more instances of the pre-training templates with their hyper-parameters, where \textbf{\textit{each template is a self-supervised learning method}}. \texttt{UniTS} has implemented various types of template, as discussed in Section~\ref{sec:pre-training-module}. During pre-training, each instance \textbf{\textit{separately}} learns its encoder $h_m$ ($m = 1, \ldots, M$), while all encoders $h_1, \ldots, h_M$ are \textbf{\textit{jointly}} used for the analysis tasks. The variety of pre-trained representations can be complementary with each other to achieve better performance.

After pre-training, the feature fusion module combines all learned representations of each sample $\xv_i$ into one embedding, denoted as $\zv_i^\prime \in \mathbb{R}^{K^\prime}$. The goal of this module is to \textbf{\textit{automatically}} fuse the information from different pre-training instances to better facilitate the tasks. The details of this module are shown in Section~\ref{sec:feature-fusion}.

Any analysis task $\mathcal{T}$ can be performed on top of $\zv_i^\prime$ by using a \textbf{\textit{task-specific}} function $g_{_{\mathcal{T}}}$ to map $\zv_i^\prime$ to the prediction $\hat{\yv}_i$ (or $\hat{y}_i$) of the target, and then \textbf{\textit{fine-tuning}} the models (including the encoders, the learnable feature fusion model and the task-specific layers) by minimizing a loss function of the task. Presently, \texttt{UniTS} supports the five mainstream tasks illustrated in Table~\ref{tab:task_examples}, while other tasks can be seamlessly integrated using the sklearn-style APIs. The analysis task module is discussed in detail in Section~\ref{sec:tasks}.

\textbf{\underline{Discussion}.} From the above description, \textit{any time series analysis task can be carried out with \texttt{ UniTS} in a \textbf{\textit{unified}} way}. At the training stage, the model $f_{_{\mathcal{T}}}$ is built following the above pipeline. During the inference stage, the learned $f_{_{\mathcal{T}}}$ is used to map the input series to the predictions. To deal with the problems of partial labeling, only a small size of the labeled data is required for fine-tuning, while \textbf{\textit{the pre-training stage does not rely on any labels (Figure~\ref{fig:sub:pipe}).}} Similarly, when facing the problem of domain shift, \textbf{\textit{the user can pre-train the encoders using the data from an available source domain to get the transferable representations}}, and then fine-tune $f_{_{\mathcal{T}}}$ using a small data set from the target domain (Figure~\ref{fig:sub:pipe}). In any case, \textbf{\textit{the pre-trained encoders can be directly used without re-training}}, and the data size and the number of iterations for fine-tuning can be much smaller than the training from scratch~\cite{CSL} while guaranteeing competitive performance (Figure~\ref{fig:performance}).      


\vspace{-0.5ex}
\section{System Internals}
This section introduces the technical details of the three main modules in \texttt{UniTS}. \textbf{\textit{All the techniques mentioned below have been implemented in the current version.}}

\vspace{-1.2ex}
\subsection{Pre-training Module}\label{sec:pre-training-module}
\texttt{UniTS} provides various self-supervised representation learning methods that can be used as pre-training templates.  Below, we explain the currently implemented pre-training templates, and discuss how to easily add new templates via the sklearn-like APIs.

In general, \texttt{UniTS} has integrated the following pre-training methods, which are \textbf{\textit{divided into three types based on different pre-training objective functions}} (see the references for more details).

\textbf{\underline{Contrastive Learning}.}  \texttt{UniTS} currently supports time series contrastive learning at three levels, i.e. the whole-series level~\cite{CSL}, the sub-sequence level~\cite{T-Loss}, and the timestamp level~\cite{ts2vec}. These methods \textbf{\textit{encourage the representations generated from the similar inputs to be closer than those of dissimilar inputs}}, which has shown effective in extracting informative features.

\textbf{\underline{Autoregression.}} The autoregression-based pre-training algorithm~\cite{TST} learns the representations by masking some observations of the input series (e.g., set to 0) and then \textbf{\textit{predicting the masked values}} using the unmasked data.  It is inspired by the masked language model in natural language processing, since both time series and sentences share the same nature of sequential dependencies.

\textbf{\underline{Hybrid}.}  This branch of approach~\cite{TS-TCC} optimizes a \textbf{\textit{hybrid objective of the two types above}}, which may be better than the individual objectives in some cases, but not always.

The pre-training template is designed using a unified sklearn-like interface to allow flexible extension. A new algorithm can be seamlessly integrated as long as it is wrapped in a Python class with two methods: \textit{fit} which takes the input $\Xv$ for pre-training, and \textit{transform} which maps $\Xv$ to the representations $\Zv$.   

\vspace{-1.5ex}
\subsection{Feature Fusion Module}\label{sec:feature-fusion}
This module fuses the representations of a variety of pre-training instances. The module is also designed using sklearn-style interfaces. A feature fusion class contains a \textit{transform} function that maps a set of representations $(\Zv_1,\ldots,\Zv_M)$ into one unified $\Zv^{\prime} \in \mathbb{R}^{N \times K^\prime}$, and $\Zv^{\prime}$ is used for the analysis tasks. Note that the parameters of a feature fusion instance can be optimized at the fine-tuning stage if they are learnable, which allows \textbf{\textit{automatically extracting information from all pre-trained representations}}. \texttt{UniTS} now supports two basic feature fusion methods as described below.      

\textbf{\underline{Concatenation.}} The simplest but most popular way of feature fusion is directly concatenating the features of each sample, i.e., $\zv_i^\prime = \zv_{i,1} \oplus \ldots \oplus \zv_{i,M}$, where $\oplus$ is the concatenation operation.

\textbf{\underline{Projection.}} One can also use a learnable model $p$ to project the concatenated features into another latent space. This is especially effective in some cases such as clustering where dimension reduction is usually required. Formally, we have $\zv_i^\prime = p(\zv_{i,1} \oplus \ldots \oplus \zv_{i,M})$.

\vspace{-2.8ex}
\subsection{Analysis Task Module}\label{sec:tasks}
The analysis task module is designed as a template wrapped with the sklearn-like APIs, where each task is an instance. The template contains two major components: a \textit{fit} method which performs the task-specific fine-tuning, and a \textit{predict} method which outputs the final prediction. \texttt{UniTS} currently supports the five important tasks described in Table~\ref{tab:task_examples}. Below, we briefly explain the technical details.

\textbf{\underline{Classification.}} This task is performed in a \textbf{\textit{standard}} manner. The output model projects the representations into $C$ classes, and the \textbf{\textit{softmax}} function is used to generate the class distribution. The standard \textbf{\textit{cross-entropy loss}} is used for fine-tuning, and the class that gives the maximum probability is determined as the prediction.

\textbf{\underline{Clustering.}} The clustering task can be directly performed by running a classical clustering algorithm (e.g., $k$-Means) on top of the representations. Moreover, one can also fine-tune the models for better performance. The fine-tuning is designed based on the \textbf{\textit{$k$-Means loss}}. At each epoch, the $k$-Means algorithm is run on the representations to obtain the $C$ centroids. Then, the sum of the L2 norms between each representation vector and its centroid is added to the pre-training loss as a regularization term, which encourages the clustering structure of the representations. Note that we do not directly minimize the k-means loss to avoid the trivial solution, i.e, all representations become equal to their centroids.

\textbf{\underline{Forecasting.}} Forecasting is performed in the \textbf{\textit{standard}} way. A decoder is used to transform the representations into predictions. Then, the forecasting loss such as the \textbf{\textit{mean square error}} (MSE) or the \textbf{\textit{mean absolute error}} (MAE) is minimized for fine-tuning. In the inference stage, the decoder outputs are the forecasted values. 



\textbf{\underline{Anomaly detection.}} We employ a popular \textbf{\textit{reconstruction-based}} framework~\cite{vldb22-ad-evaluation} for anomaly detection. A decoder is added to project the representations to the predicted inputs $\hat{\xv}_i$. The objective is to minimize the \textbf{\textit{reconstruction loss}} as $||\xv_i - \hat{\xv}_i||$. For detection, an anomaly score $s_t$ is computed at each time step $t$ as $|\hat{\xv}_{i,t} - \xv_{i,t}|$. The observation with the score larger than a threshold $\tau$ is determined as an anomaly, i.e., $\hat{\yv}_{i,t} = 1$ if $s_t > \tau$ and 0 otherwise.

\textbf{\underline{Missing value imputation.}} This task is performed using a modern structure named \textbf{\textit{denoising autoencoder}} (DAE)~\cite{dae}. During fine-tuning, we generate a random binary mask $\boldsymbol{m}_i \in \{0,1\}^{D \times T}$ for each $\xv_i$. The masked sample $\xv_i \otimes \boldsymbol{m}_i$ is input to the pre-trained encoders, where $\otimes$ denotes the element-wise multiplication. Then, a decoder is used on top of the representations to reconstruct the entire input $\xv_i$ as $\hat{\xv}_i$. DAE is learned by minimizing the \textbf{\textit{reconstruction loss}} $||\xv_i - \hat{\xv}_i||$. At the inference stage, the missing values of an input $\xv_i$ are first replaced by 0, i.e.,  $\xv_{i,j,k} = 0, \forall (j,k) \in P_i$. Next, $\xv_i$ is input into the fine-tuned model to predict $\hat{\xv}_i$. To this end, the missing value at position $(j,k) \in P_i$ is imputed using $\hat{\xv}_{i,j,k}$.

\begin{figure}[t]
	\centering
	\subcaptionbox{ Pipelines for partially-labeled and cross-domain tasks.\label{fig:sub:pipe}}
	{\includegraphics[width=.51\linewidth]{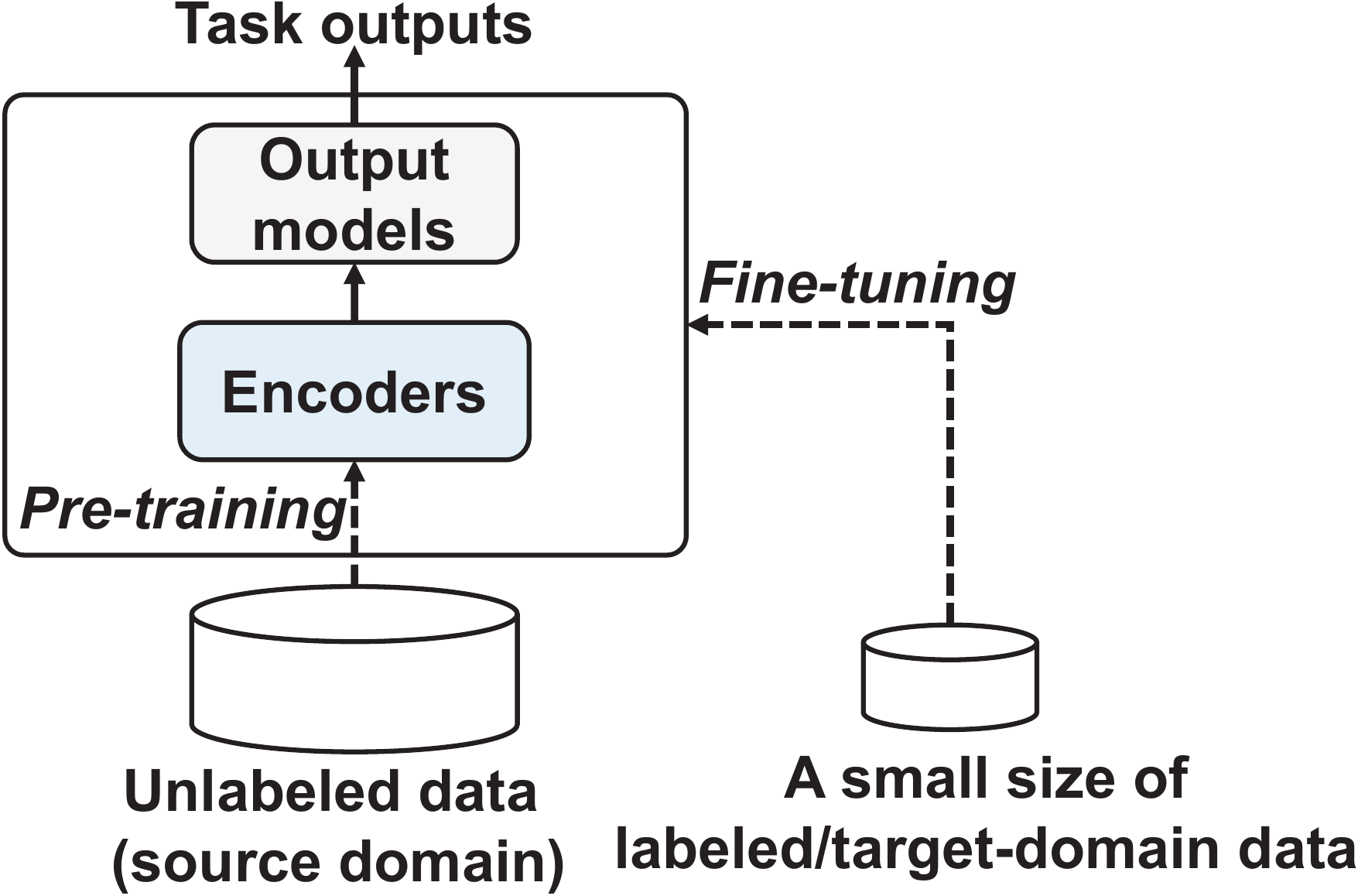}}
	\subcaptionbox{\centering \texttt{UniTS} interfaces.\label{fig:sub:gui}}
	{\includegraphics[width=.48\linewidth]{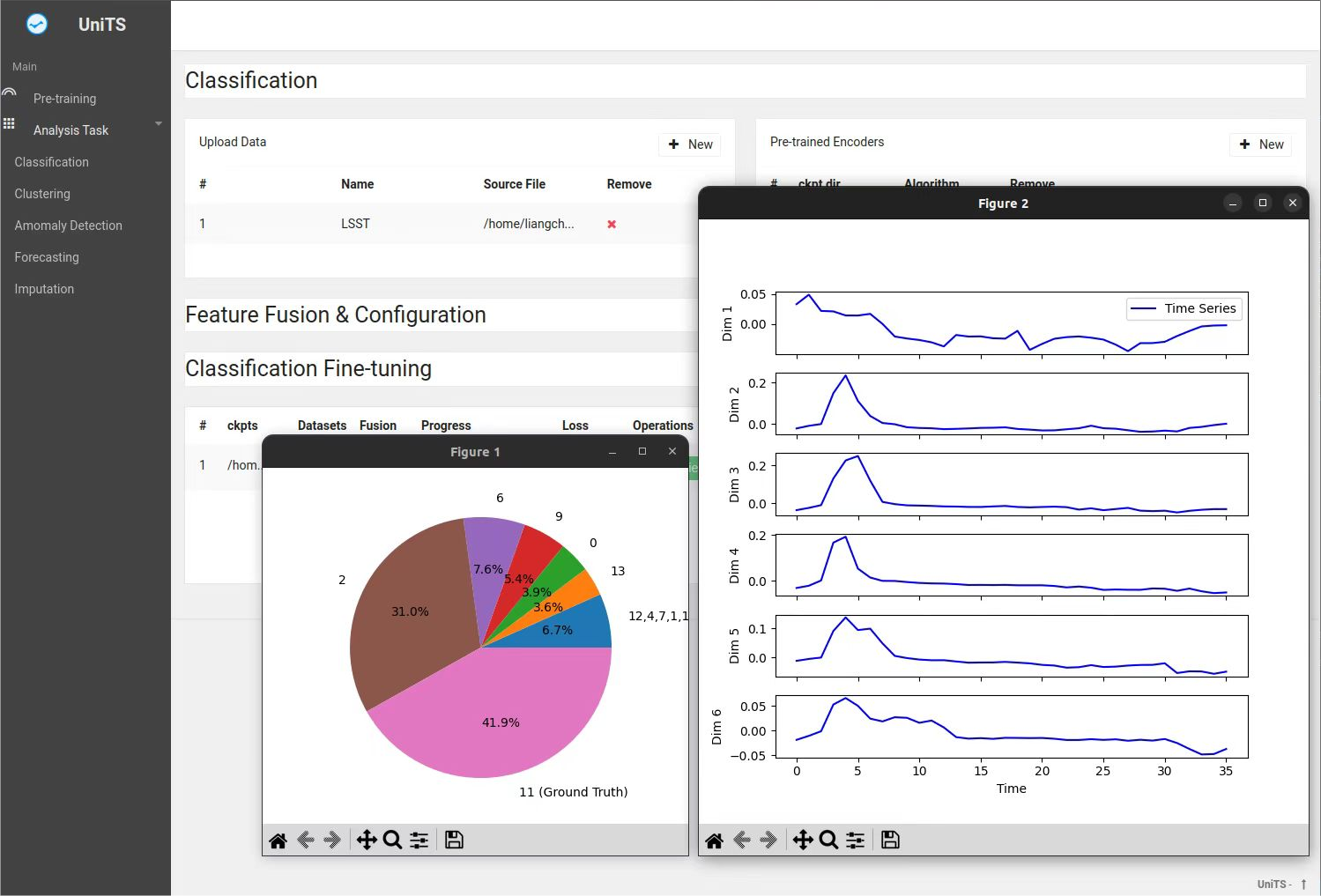}}
 \vspace{-1.5ex}
	\caption{Demonstration overview of \texttt{UniTS}}
  \label{fig:demo}
\end{figure}

\section{Demonstration}
In our demonstration, we intend to show how \texttt{UniTS} can help the user to perform different time series analysis tasks, and how it can tackle the practical problems of partial labeling and domain shift. 

\textbf{\underline{Unified pipeline for time series analysis.}} We prepare the UEA datasets~\cite{UEA} for the audiences to interact with \texttt{UniTS}. They can also analyze their own data using our system. Overall, the user takes the following two steps to perform an anslysis task using \texttt{UniTS}. 

\textbf{\text{$\ \ \bullet$ Pre-training.}} In this step, the user adopts the \texttt{UniTS} interfaces to load the data set and configure the pre-training methods with the templates to generate the instances, and then clicks the ``Pre-training'' button to start the self-supervised learning. \texttt{UniTS} visualizes the loss curves to help the user for monitoring. The user can select and save the encoders for the analysis tasks. \textit{Note that this step is \textbf{only needed once}. All the following processes can be repeatedly performed using the pre-trained encoders without re-training.}    

\textbf{\text{$\ \ \bullet$ Fine-tuning.}} In this stage, the user loads the pre-trained encoders and the training data of the task. The user can configure the feature fusion and analysis task modules through the GUIs. Once the ``Fine-tuning'' button is activated, \texttt{UniTS} starts to learn the model for the selected task. The loss curves are plotted as the pre-training process. \texttt{UniTS} also provides the visualization and evaluation of the results for different tasks to validate the models.  After fine-tuning, the user can save the model as a standard JSON file which can be employed by any machine learning tool for inference.

\textbf{\underline{Addressing the practical problems.}} 
Benefiting from the self-supervised pre-training which learns informative and transferable features, the user can  tackle the problems of partial labeling and domain shift using \texttt{UniTS}. The process is illustrated in Figure~\ref{fig:sub:pipe}.

\begin{figure}[t]
    \centering
    \includegraphics[width=.88\linewidth]{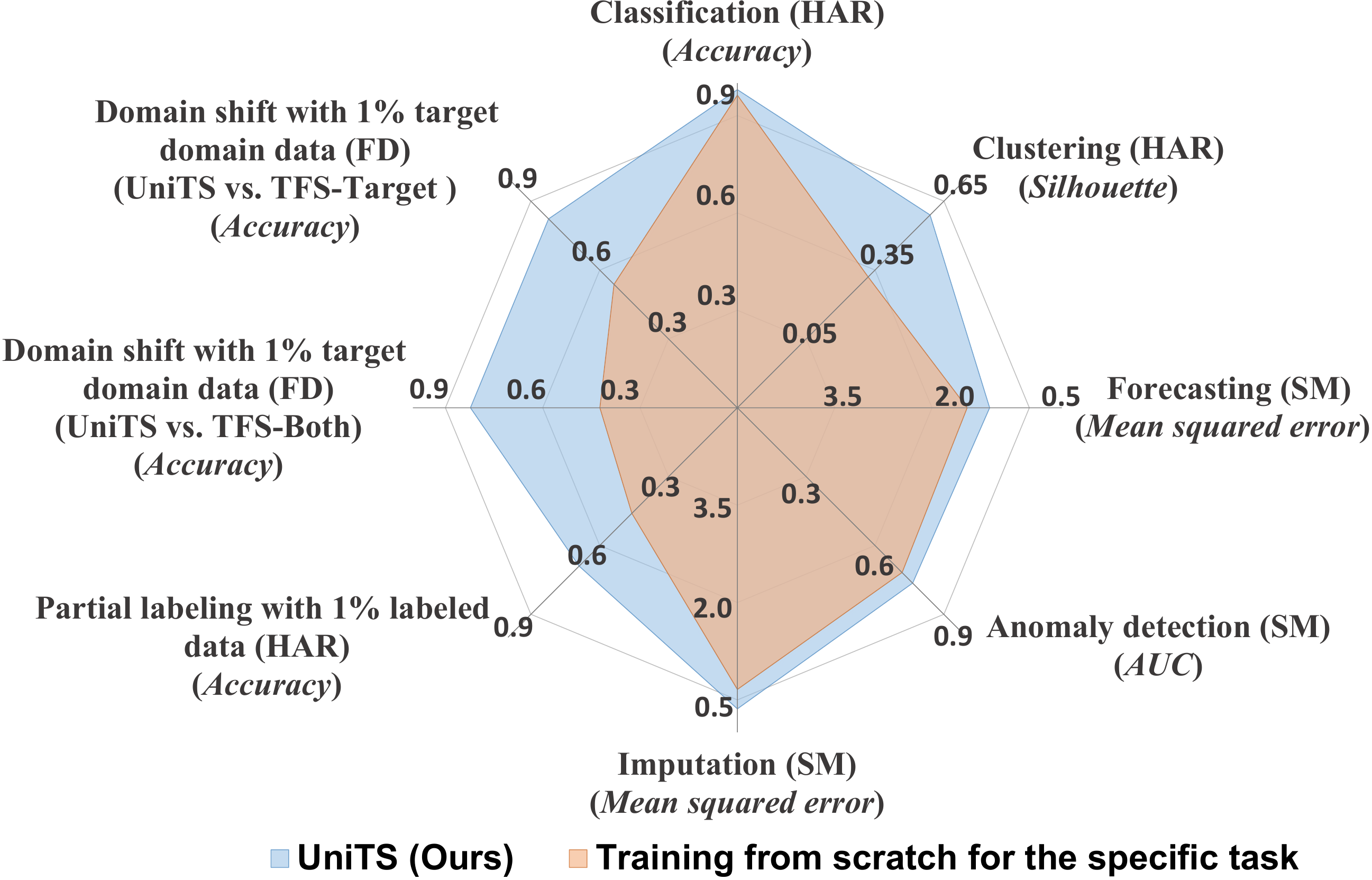}
    \vspace{-1.6ex}
    \caption{Overall performance of \texttt{UniTS}}
    \label{fig:performance}
    \vspace{-0.6ex}
\end{figure}

\text{\textbf{$\ \ \bullet$ Partial labeling.}} In this scenario, the user only needs to load the available labeled data for fine-tuning. \texttt{UniTS} can achieve \textbf{\textit{significantly better performance}} compared to traditional training from scratch using only the labeled data (Figure~\ref{fig:performance}).

\text{\textbf{$\ \ \bullet$ Domain shift.}} To achieve cross-domain analysis using \texttt{UniTS}, the user can fine-tune the models with only a small size of data from the target domain based on the pre-trained encoders. The models learned with \texttt{UniTS} can be \textbf{\textit{more generalizable}} than the models \underline{\textbf{t}}rained \underline{\textbf{f}}rom \underline{\textbf{s}}cratch using the same data from the target domain (i.e. \texttt{TFS-Target}) or both domains (\texttt{TFS-Both}) (Figure~\ref{fig:performance}).

\textbf{\underline{Superior performance.}} We also evaluate \texttt{UniTS} using real-world tasks and settings. Our application scenarios include \textbf{\textit{human action recognition (HAR)}},  \textbf{\textit{fault detection (FD)}} for two machines under different working conditions (a.k.a. domains) and \textbf{\textit{server monitoring (SM)}}. For benchmarking, we create one instance for every pre-training template described in Sec.~\ref{sec:pre-training-module} using their \textit{default encoders}, and use a \textit{linear layer} to fuse all representations into 64 dimensions. For task-specific output models, we use a standard \textit{linear layer for classification tasks} and a simple \textit{multilayer perceptron (MLP) with one hidden layer as the decoder} for forecasting, anomaly detection, and imputation. We compare \texttt{UniTS} with the traditional solutions of \textit{directly training the entire model $f_{_{\mathcal{T}}}$ (encoders + fusion layer + output model) from scratch using the task-specific loss function without self-supervised pre-training.} The model architectures are the same as \texttt{UniTS}.  The results in Figure~\ref{fig:performance} indicate the \textbf{\textit{superiority of \texttt{UniTS} powered by self-supervised representation learning.}}

\begin{acks}
 This paper was supported by the NSFC grant (62232005, 62202126), the National Key Research and Development Program of China (2021YFB3300502), and the Postdoctoral Fellowship Program of CPSF (GZC20233457).
\end{acks}
\normalem
\bibliographystyle{ACM-Reference-Format}
\bibliography{sample}
\end{document}